\PassOptionsToPackage{table,dvipsnames}{xcolor}
\documentclass[]{selfevolagent}
\usepackage{microtype}
\usepackage{amsfonts}
\usepackage{xcolor}
\definecolor{lavender}{RGB}{230,230,250} % 淡紫色
\definecolor{softlavender}{RGB}{238, 223, 255} % 你原来用的 softlavender
\definecolor{selfevolagent}{RGB}{220,211,237} % survey的颜色
\usepackage{wrapfig}
\usepackage{float}
\usepackage{amsmath}
\usepackage{amsthm}
\usepackage{subcaption}
\usepackage{makecell}
\usepackage{algpseudocode}
\usepackage[linesnumbered,lined,boxed,commentsnumbered,ruled,longend]{algorithm2e}
\theoremstyle{plain}

\theoremstyle{definition}

\theoremstyle{remark}

\usepackage{enumitem}
\usepackage{pifont}
\usepackage[edges]{forest}
\usetikzlibrary{arrows.meta}
\usepackage{tikz}
\usepackage{graphicx}     % for \resizebox
\usepackage{booktabs}     % for \toprule, \midrule, \bottomrule
\usepackage{enumitem}     % to use [nosep,leftmargin=*] with itemize
\usepackage[most]{tcolorbox} % 画彩色框
\usepackage{url}
\usepackage{tabularx}
\usepackage{fontawesome5}
\usepackage{svg}
\usepackage{xltabular}
\usepackage{caption}

\usetikzlibrary{
  positioning,
  calc,
  shapes.symbols,
  shapes.geometric,
  shapes.misc
}
\usepackage[T1]{fontenc}
\definecolor{selfevolagent_dark}{HTML}{37D2A6} 
\definecolor{selfevolagent_light}{HTML}{9BE9D3}
\definecolor{selfevolagent_lighter}{HTML}{CDF4E9}

\newcommand{\ghlink}[1]{\faIcon{github}\,\href{#1}{GitHub}}
\newcommand{\weblink}[1]{\faIcon{globe}\,\href{#1}{Website}}

\newcommand{\ie}{\textit{i.e.}}
\newcommand{\eg}{\textit{e.g.}}
\newcommand{\ours}{\texttt{GISA}}

\definecolor{rootcolor}{RGB}{101, 45, 144}   % 紫色
\definecolor{catcolor}{RGB}{255, 192, 0}     % 金黄色
\definecolor{subcatcolor}{RGB}{237, 125, 49} % 橙色
\definecolor{papercolor}{RGB}{68, 114, 196}  % 蓝色

\definecolor{prompttitle}{HTML}{2896FD}
\definecolor{promptbg}{HTML}{EDF2FB}
\newtcolorbox{promptbox}[1][]{
  colback=promptbg,      % 背景颜色：浅灰
  colframe=prompttitle,    % 边框颜色：深灰
  coltitle=white,      % 标题颜色
  boxrule=0.5pt,       % 边框粗细
  arc=2pt,             % 圆角
  left=5pt, right=5pt, top=5pt, bottom=5pt,
  fontupper=\small, % 内容字体：等宽字体，小号
  title=\textbf{System Prompt for ReAct-based Agent}, % 盒子标题
  #1
}
\newtcolorbox{toolbox}[1][]{
  colback=promptbg,
  colframe=prompttitle,
  coltitle=white,
  boxrule=0.5pt,
  arc=2pt,
  left=5pt, right=5pt, top=5pt, bottom=5pt,
  fontupper=\small, % 这里不用全篇等宽字体，混合使用更清晰
  title=\textbf{Tools Definition},
  #1
}

\newcommand{\param}[2]{\texttt{\textbf{#1}} \textit{(#2)}}

\title{GISA: A Benchmark for General Information Seeking Assistant}
\author{Yutao Zhu$^{1,\dagger,\text{\faIcon[regular]{envelope}}}$}
\author{Xingshuo Zhang$^{1,\dagger}$}
\author{Maosen Zhang$^{1,\dagger}$}
\author{Jiajie Jin$^1$}
\author{Liancheng Zhang$^1$}
\author{Xiaoshuai Song$^1$}
\author{Kangzhi Zhao$^2$}
\author{Wencong Zeng$^2$}
\author{Ruiming Tang$^2$}
\author{Han Li$^2$}
\author{Ji-Rong Wen$^1$}
\author{Zhicheng Dou$^{1,\text{\faIcon[regular]{envelope}}}$}

\affiliation{%
\vspace{0.5em}%
{$^\dagger$Equal Contribution}

{\textbf{Affiliation}: $^1$Renmin University of China}; $^2$Kuaishou Technology}

\abstract{The advancement of large language models (LLMs) has significantly accelerated the development of search agents capable of autonomously gathering information through multi-turn web interactions. Various benchmarks have been proposed to evaluate such agents. However, existing benchmarks often construct queries backward from answers, producing unnatural tasks misaligned with real-world needs. Moreover, these benchmarks tend to focus on either locating specific information or aggregating information from multiple sources, while relying on static answer sets prone to data contamination. To bridge these gaps, we introduce GISA, a benchmark for General Information-Seeking Assistants comprising 373 human-crafted queries that reflect authentic information-seeking scenarios. GISA features four structured answer formats (item, set, list, and table), enabling deterministic evaluation. It integrates both deep reasoning and broad information aggregation within unified tasks, and includes a live subset with periodically updated answers to resist memorization. Notably, GISA provides complete human search trajectories for every query, offering gold-standard references for process-level supervision and imitation learning. Experiments on mainstream LLMs and commercial search products reveal that even the best-performing model achieves only 19.30\% exact match score, with performance notably degrading on tasks requiring complex planning and comprehensive information gathering. These findings highlight substantial room for future improvement.}
\metadata[\faEnvelope\ Contact]{yutaozhu94@gmail.com, dou@ruc.edu.cn}
\metadata[{\raisebox{-0.2ex}{\includegraphics[height=1.1em]{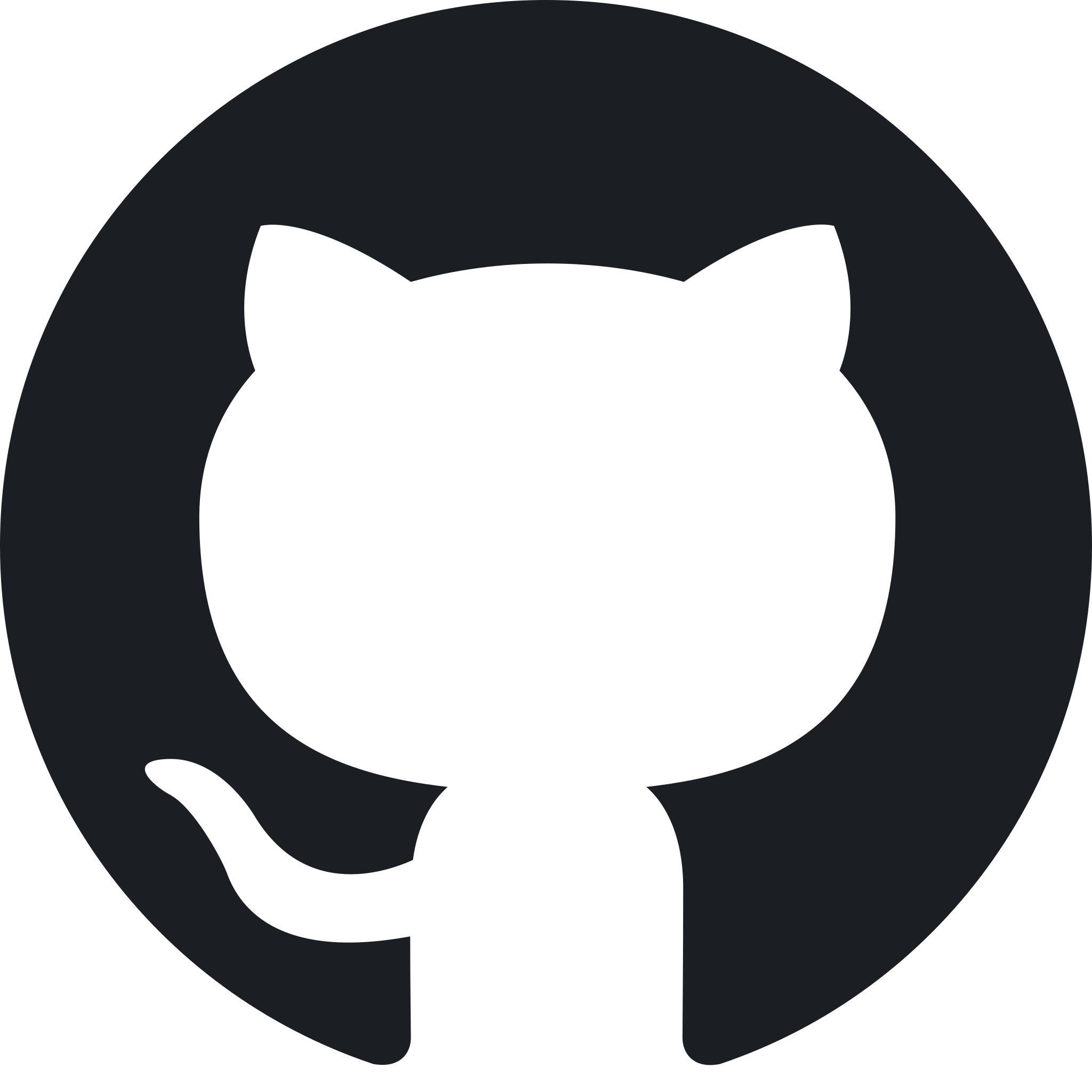}}\ Code}]{\url{https://github.com/RUC-NLPIR/GISA}}
\metadata[\faChartLine\ Leaderboard]{\url{https://huggingface.co/spaces/RUC-NLPIR/GISA-LeaderBoard}}

\begin{document}

\maketitle

\section{Introduction}
With the development of large language models (LLMs), LLM-driven agents have been widely applied in many tasks~\citep{DBLP:conf/uist/ParkOCMLB23,DBLP:conf/nips/YuYLDJCCSCLXZSX24,DBLP:conf/acl/QianLLCDL0CSCXL24,WebThinker,DBLP:journals/corr/abs-2404-02831}. Among these applications, information retrieval stands out as a fundamental need in daily life that has been notably transformed by agent-based systems. Traditional search engines require users to manually formulate queries, navigate through multiple web pages, and synthesize information themselves. This process is often time-consuming and requires significant effort. In contrast, search agents leverage the strong reasoning capabilities of modern LLMs to understand complex user information needs. These agents can autonomously conduct multiple rounds of searching and web browsing to collect relevant information and fulfill user requirements. By automating the process of querying and information gathering, search agents significantly reduce user burden and improve information access efficiency. As a critical tool for information acquisition, search agents have attracted increasing attention from both academia and industry~\citep{Search-o1,WebSailor,WebThinker}, becoming essential components in many research-oriented products and knowledge discovery platforms (such as OpenAI Deep Research and Gemini Deep Research).\footnote{OpenAI Deep Research: \url{https://openai.com/index/introducing-deep-research/}, Gemini Deep Research: \url{https://gemini.google/overview/deep-research/}.}

\begin{table*}[t]
    \centering
    \small
    \caption{Comparison between \ours{} with other agentic search benchmarks.}
    \setlength{\tabcolsep}{1.9mm}{
    \begin{tabular}{lccccccccc}
    \toprule
    \textbf{Benchmark} & \textbf{Deep} & \textbf{Wide} & \textbf{Question-driven} & \textbf{Human Trajectory} & \textbf{Live} & \textbf{Answer Type} & \textbf{Evaluation} \\
    \midrule
    BrowseComp-EN/ZH & $\checkmark$ & $\times$ & $\times$  & $\times$ & $\times$ & Item & LLM \\
    InfoDeepSeek& $\checkmark$ & $\times$ & $\times$  & $\times$ & $\times$ & Item & LLM \\
    ScholarSearch & $\checkmark$ & $\times$ & $\checkmark$ & $\times$ & $\times$ & Item & LLM \\
    GAIA & $\checkmark$ & $\times$ & $\checkmark$ & $\times$ & $\times$ & Item, List & Metric-based \\
    Xbench-DeepSearch& $\checkmark$ & $\times$ & Part of & $\checkmark$ & $\checkmark$ & Item & LLM \\
    % DeepResearch Bench~\citep{} & $\checkmark$ & $\checkmark$ & $\checkmark$ & $\times$ & $\times$ & Report & LLM \\
    WideSearch & $\times$ & $\checkmark$ & $\checkmark$ & $\times$ & $\times$ & Table & LLM \\
    DeepWideSearch & $\checkmark$ & $\checkmark$ & Part of & $\times$ & $\times$ & Table & LLM \\
    \midrule
    \rowcolor[RGB]{235,245,250}
    \ours{} (Ours) & $\checkmark$ & $\checkmark$ & $\checkmark$ & $\checkmark$ & $\checkmark$ & Item, Set, List, Table  & Metric-based \\
    \bottomrule
    \end{tabular}
    }
    \label{tab:comparison}
\end{table*}

A high-quality benchmark is essential to systematically evaluate the capabilities of search agents and identify directions for further improvement. During the past few years, the research community has proposed various benchmarks to evaluate agents' abilities to retrieve and synthesize information from the web~\citep{browsecomp,InfoDeepSeek,xbench,GAIA,widesearch}. However, upon closer examination, we identify several critical limitations in existing benchmarks that hinder their effectiveness in comprehensively evaluating general-purpose information-seeking agents. 
\textbf{First}, to increase task complexity or assess agents' planning capabilities over challenging problems, many existing benchmarks adopt a \textit{reverse-engineering} approach that constructs queries backward from pre-selected answers (\eg{}, BrowseComp~\citep{browsecomp}). While this approach can produce difficult tasks, the resulting queries often deviate significantly from authentic human information needs. Some constructed problems may even be unsolvable through a natural forward search process. Consequently, optimizing agent performance on such benchmarks may not translate into improved real-world user experience. 
\textbf{Second}, existing benchmarks mainly focus on evaluating agents' ability to perform deep search within the web, \ie{}, navigating through multiple links and pages to locate a specific piece of information~\citep{browsecomp,webwalker,InfoDeepSeek}. However, real-world information needs often require not only depth but also breadth, where agents must gather and aggregate information from diverse sources. Although recent efforts such as WideSearch~\citep{widesearch} have begun to explore the evaluation of broad information collection capabilities, achieving a balanced and unified evaluation of both deep and wide search remains an open challenge. 
\textbf{Third}, search serves as a primary means for humans to acquire information, particularly timely and up-to-date knowledge. Ideally, benchmarks should evolve alongside the information landscape to remain relevant. Nevertheless, most existing benchmarks, for the sake of evaluation convenience and reproducibility, rely on questions with long-term stable answers~\citep{InfoDeepSeek}. As LLMs are continuously trained on increasingly recent data, such static benchmarks may become out-of-date, as models may have already memorized the answers during pre-training, thereby failing to genuinely test their search capabilities.

To address these limitations, we propose \textbf{\ours{}}, a benchmark for \textbf{G}eneral \textbf{I}nformation-\textbf{S}eeking \textbf{A}ssistants containing carefully crafted queries along with their answers and human-annotated search trajectories. 
% All queries are designed to be challenging yet admit unique, factual answers, enabling simple and robust automatic evaluation. 
\ours{} distinguishes itself through the following key features:

\begin{itemize}[topsep=0pt,leftmargin=2em]
    \item \textbf{Diverse answer formats with deterministic evaluation}: \ours{} formulates answers into four structured types: items, sets, lists, and tables. This design enables deterministic and reproducible evaluation using strict matching metrics, avoiding the subjectivity and instability of LLM-based judgment while preserving task complexity and diversity.
    \item \textbf{Unification of deep and wide search capabilities}: Real-world information seeking often requires both exploring deep reasoning paths and aggregating information across diverse sources. \ours{} integrates both dimensions, evaluating an agent's ability to perform vertical investigation and horizontal summarization within complex, long-horizon tasks.
    \item \textbf{Dynamic and anti-static evaluation}: To prevent data contamination from pre-training memorization, \ours{} categorizes queries into stable and live subsets. The live subset requires accessing real-time information and is periodically updated, ensuring the benchmark remains challenging and resistant to memorization over time.
    \item \textbf{Process-level supervision via human trajectories}: Beyond question-answer pairs, \ours{} provides complete human search trajectories for every query. These trajectories serve as gold standards for process reward modeling and imitation learning, while also verifying that all tasks are solvable through realistic search behaviors.
\end{itemize}

% 增加构建过程以及bench的整体数据描述
We construct \ours{} through a rigorous multi-stage annotation process involving question design, answer labeling, and human trajectory collection, with each stage undergoing multiple rounds of quality verification. On average, each query requires over one hour of human effort from initial design to final validation. The final benchmark comprises 373 high-quality queries spanning diverse domains and complexity levels. To facilitate future research, we release comprehensive documentation, evaluation scripts, and a public leaderboard open for submission.

We evaluate a diverse set of state-of-the-art LLMs and commercial search products on \ours{}. 
Results show that even the best-performing model achieves only 19.30\% overall exact match, with commercial deep research systems struggling to outperform LLM-based agents.
% Results show that current systems still fall substantially short of human performance on complex information-seeking tasks. Even the best-performing model achieves only 18.23\% overall exact match
% , with performance degrading notably as tasks require broader information gathering and more structured answer organization. 
% Commercial deep research systems, despite their sophisticated designs, do not outperform LLM-based agents and often struggle with instruction following. 
These findings underscore the challenging nature of \ours{} and highlight significant opportunities for advancing general-purpose information-seeking agents.

\section{Related Work}
\textbf{Search Agent.}
LLM-driven search agents have evolved beyond early retrieval-augmented generation into complex systems capable of autonomous planning and dynamic decision-making~\citep{LLMforIRSurvey,DeepSearchAgentSurvey}.
To meet diverse information needs, current research employs structures ranging from breadth-oriented parallel search to depth-oriented serial reasoning~\citep{Search-o1,HybridDeepSearcher,ParallelMuse}, while further integrating complex architectures like knowledge graph, Monte Carlo Tree Search (MCTS), and multi-agent collaboration~\citep{DeepDive,HG-MCTS,ManuSearch}.
In terms of optimization, combining Supervised Fine-Tuning (SFT) and Reinforcement Learning (RL) significantly enhances agent planning and tool usage in dynamic environments, and drives the emergence of applications centered on web search and research~\citep{AgenticSearchRLSurvey,WebThinker,Search-R1,WebSailor}.
Despite these advances, current evaluation systems remain limited by unrealistic task construction, a lack of unified assessment for deep reasoning and breadth aggregation, over-reliance on unstable LLM-based scoring, and data contamination in static benchmarks. 
Consequently, accurately measuring comprehensive agent performance in real-world, dynamic web environments remains a challenge.

\noindent\textbf{Benchmark for Agentic Search.}
The rapid advancement of search agents has given rise to a variety of benchmarks, as shown in Table~\ref{tab:comparison}.
Benchmarks including BrowseComp~\citep{browsecomp}, InfoDeepSeek~\citep{InfoDeepSeek}, and xbench-DeepSearch~\citep{xbench} focus on deep search, evaluating multi-hop reasoning and cross-page information integration~\citep{GAIA,RAVine,SealQA,xbench,BrowseComp-ZH,LiveSearchBench,ScholarSearch}. However, tasks in these datasets are often reverse-engineered from answers; while challenging, they may deviate from real user queries.
Conversely, benchmarks like WideSearch~\citep{widesearch} emphasize reliability in large-scale atomic information collection and structured organization. 
% Yet, their tasks are often highly mechanical, failing to capture the complexity of human information-seeking patterns.
While DeepWideSearch~\citep{DeepWideSearch} attempts to evaluate both depth and breadth by combining BrowseComp and WideSearch, it still lacks task-driven construction and the support of real user behavioral trajectories.
Moreover, most existing benchmarks rely on static snapshots, making it difficult to discern whether an agent acts based on web information or its own internal knowledge.
In comparison, \ours{} provides a realistic, dynamic, and process‑interpretable benchmark. To ensure tasks are meaningful and aligned with actual information needs, \ours{} centers on human-designed tasks covering both deep and wide information seeking. To prevent data contamination, \ours{} maintains queries involving live and evolving information. Finally, \ours{} provides complete human search trajectories, enabling fine‑grained process analysis and serving as references for training agents to emulate human search strategies.

\begin{figure*}[t]
    \centering
    \includegraphics[width=\linewidth]{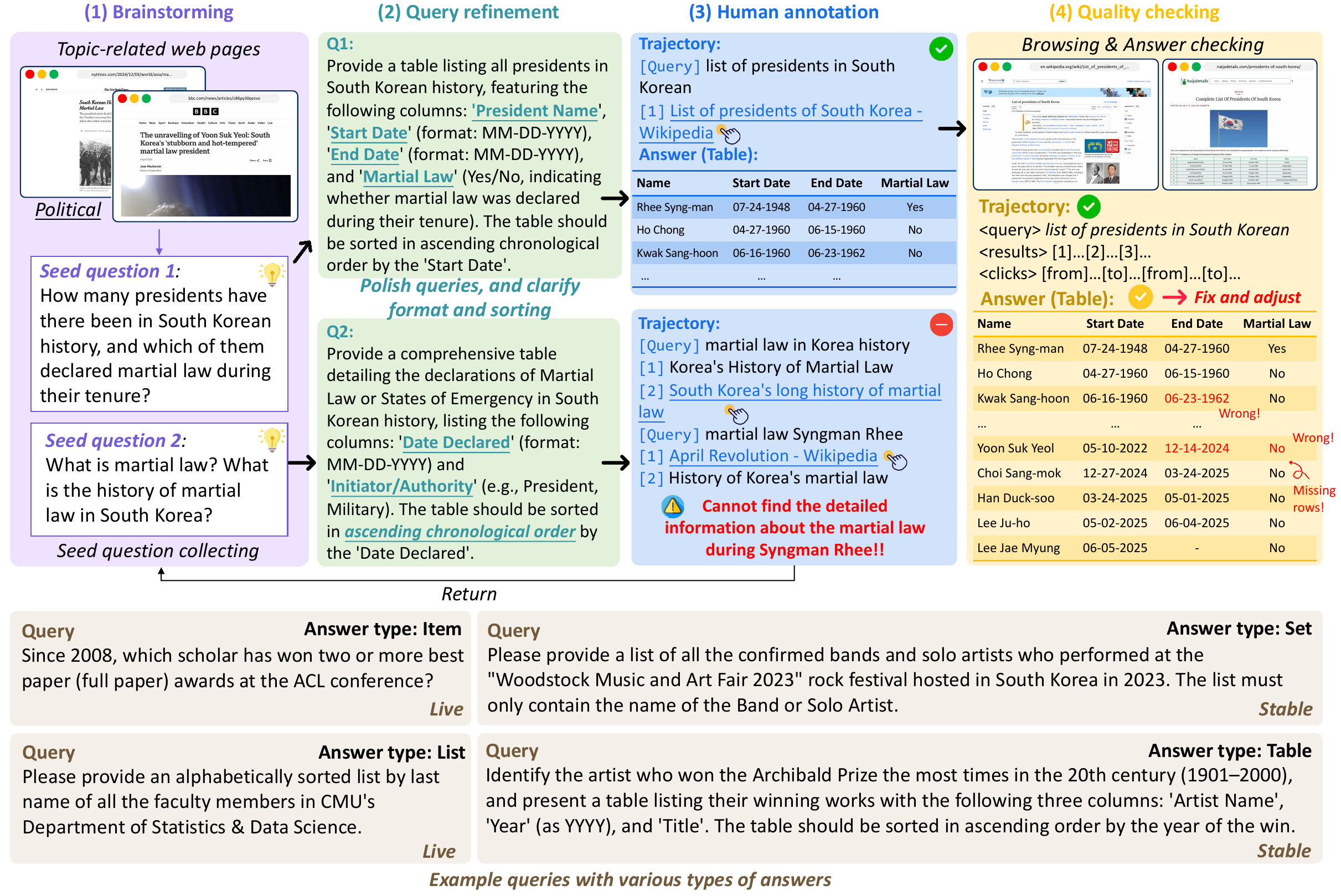}
    \caption{Illustration of benchmark construction process and some example queries with various types of answers.}
    \label{fig:method}
\end{figure*}

\begin{figure}[t]
     \centering
     \begin{minipage}[b]{0.49\linewidth}
         \centering
         \includegraphics[width=\linewidth]{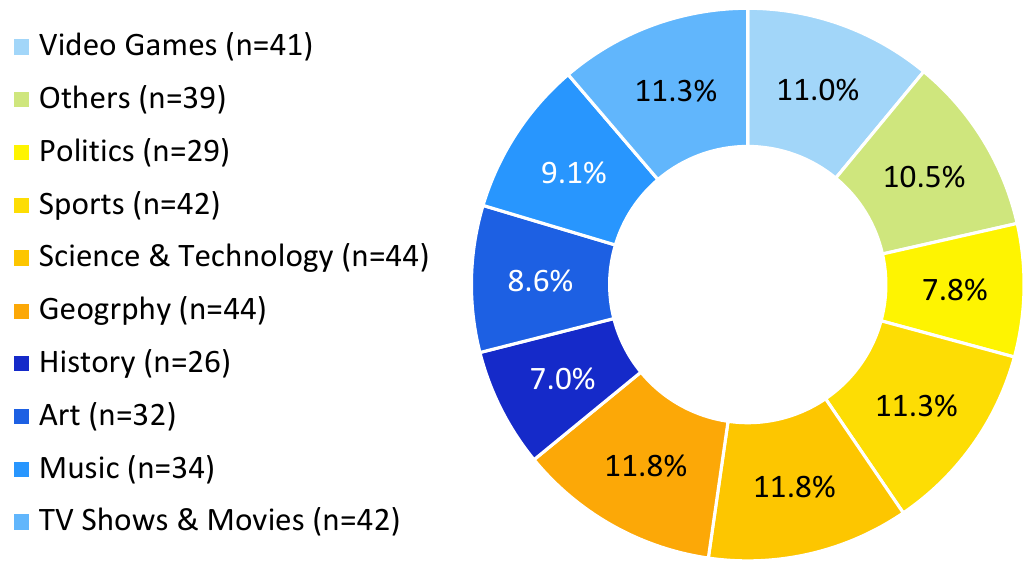}
         \caption{Distribution of topics in \ours{}.}
         \label{fig:topic}
     \end{minipage}
     \hfill
     \begin{minipage}[b]{0.45\linewidth}
         \centering
         \includegraphics[width=\linewidth]{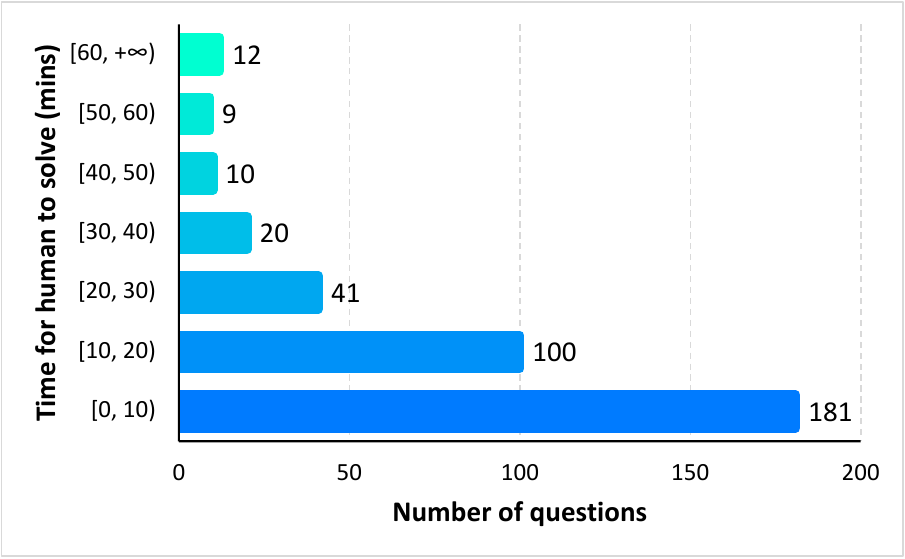}
         \caption{Statistics of human annotation time on \ours{}.}
         \label{fig:minitues}
     \end{minipage}
\end{figure}

% \begin{figure}
%     \centering
%     \includegraphics[width=.9\linewidth]{fig/topic.pdf}
%     \caption{Distribution of topics in \ours{}.}
%     \label{fig:topic}
% \end{figure}

\section{Benchmark}
In this section, we detail the design and construction process of \ours{}, followed by the evaluation protocol.

\subsection{Construction}
The construction of \ours{} is a human-centered process designed to ensure that each query is challenging, realistic, and strictly aligned with the goal of evaluating real-world information-seeking capabilities. The entire workflow, ranging from initial question design to final inclusion, is governed by rigorous criteria and multi-stage verification. As illustrated in Figure~\ref{fig:method}, the construction pipeline comprises four distinct stages: (1)~brainstorming, (2)~query refinement, (3)~human annotation, and (4)~quality checking.

\subsubsection{Brainstorming}
To ensure the benchmark covers a diverse range of topics that truly reflect human interests, we adopt the taxonomy from BrowseComp~\citep{browsecomp}. This taxonomy categorizes information needs into ten distinct domains: \textit{TV Shows \& Movies}, \textit{Science \& Technology}, \textit{Art}, \textit{History}, \textit{Sports}, \textit{Music}, \textit{Video Games}, \textit{Geography}, \textit{Politics}, and \textit{Other}. The distribution of topics in \ours{} is illustrated in Figure~\ref{fig:topic}.

Based on these categories, we employ an open-ended browsing strategy to stimulate query generation. Annotators are instructed to visit domain-specific websites, such as news aggregators for \textit{Politics} or encyclopedic archives for \textit{History}, and freely explore content. During this process, they are encouraged to document any questions that arise spontaneously from the information they consume. This approach mimics the natural cognitive process of ``encountering information and seeking further knowledge''~\citep{curiosity}.
For instance, as illustrated in Figure~\ref{fig:method}, an annotator browsing political news might encounter a report regarding President Yoon Suk Yeol and the declaration of martial law. This stimulus could trigger specific inquiries, such as \textit{What is the definition of martial law?} or \textit{Are there historical precedents for martial law in South Korea?} By systematizing this browsing-inspired workflow across all ten domains, we successfully obtain a large collection of seed questions.

\subsubsection{Query Refinement}
After obtaining a diverse pool of seed questions, the next step is to transform these raw questions into formal and structured queries. This stage is critical for realizing two core design goals of \ours{}: enabling deterministic evaluation through structured answer formats, and unifying deep and wide search capabilities within realistic tasks.

In this stage, annotators formulate the final query while determining the optimal format of the target answer. We categorize the target answer structures into four types: \textit{item}, \textit{set}, \textit{list}, and \textit{table} (example queries are shown in the bottom side of Figure~\ref{fig:method}). This structured design enables \textbf{deterministic and reproducible} evaluation using strict matching metrics, avoiding the subjectivity of LLM-based judgment. If a query is compatible with multiple formats, annotators are allowed to select the most appropriate format. Besides, to ensure the uniqueness of search trajectories in the dataset, we adopt a strict one-to-one rule: each seed question is used to construct exactly one formal query. 
% This prevents the generation of multiple queries that rely on identical or highly similar search processes.

Crucially, this formulation process involves a \textit{preliminary feasibility check} that ensures queries require genuine search effort spanning both depth and breadth. Annotators are required to perform a quick search to verify that the target answer is not readily available in a pre-aggregated form on existing web pages. For example, rather than simply asking for the \textit{``total count of South Korean presidents''}, a fact often directly displayed in search snippets, we instead request a structured table listing all presidents along with whether each declared martial law during their tenure (as illustrated in Figure~\ref{fig:method}). This task exemplifies \textbf{the integration of deep and wide search}: the agent must first gather basic information about all presidents (broad aggregation), then investigate each president's historical record to determine martial law declarations (deep reasoning for individual verification). Such queries cannot be answered by retrieving a single pre-existing source, but require systematic collection and synthesis across multiple pages. This filtering step eliminates trivial queries and ensures that \ours{} evaluates agents' ability to perform both vertical investigation and horizontal summarization within unified tasks.

Furthermore, we address the temporal nature of information needs. Annotators must classify each query as either \textit{stable} (immutable for at least three years) or \textit{live} (subject to change). For the latter, we commit to a \textbf{monthly maintenance schedule}, updating the ground-truth regularly to ensure the benchmark remains accurate over time.

Beyond general formulation, we impose strict constraints for structured output formats to ensure the evaluation is deterministic: (1) For list-type queries, the sorting criterion must be explicitly defined in the prompt (\eg{}, ``sorted alphabetically by name'') to avoid ambiguity in the output order. (2) For table-type queries, the schema must be fully specified. Annotators must define the exact column names to be retrieved and designate a primary column for sorting. To handle duplicate values in the primary column, multi-level sorting criteria (\ie{}, secondary and tertiary keys) must be provided as tie-breaking rules, applied sequentially until a unique row order is guaranteed.

\subsubsection{Human Annotation}
Prior to the official annotation, a pilot phase is conducted using five trial questions. This phase serves to train annotators on tool usage and align their outputs with our strict formatting specifications (\eg{}, ensuring column headers match the table schema, and verifying sorting logic). Feedback is provided iteratively during this phase. To maintain data integrity, these pilot samples and their ground-truths are excluded from the final dataset.

To capture the fine-grained decision-making processes of human searchers, we develope a custom browser extension. This tool operates in the background to log interaction data, including: (1) search queries issued to Google; (2) content of search engine results pages; (3) click-through behaviors; and (4) precise timestamps for duration analysis.

The annotation workflow follows a standardized protocol: 

(1) \textbf{Initialization:} The annotator activates the extension to begin a session for a specific query. 

(2) \textbf{Search and restrictions:} Annotators are restricted to using Google Search as their retrieval tool.\footnote{Given that Google Search may occasionally trigger automatic AI-generated summaries, annotators are explicitly instructed to ignore these modules. They are strictly prohibited from viewing or utilizing any information provided by these built-in LLMs. This constraint ensures that the recorded trajectories reflect a complete and organic human information-seeking process rather than the passive consumption of AI-synthesized answers.} While they may navigate freely through resulting web links, the use of other search engines or LLMs is prohibited.

(3) \textbf{Exception handling:} In cases where the target information is inaccessible, or where sources exhibit irreconcilable conflicts, annotators are instructed to document the specific issue and skip the query. These flagged queries are collected and returned to the \textit{query refinement} stage for revision. 

(4) \textbf{Answer construction:} Annotators use spreadsheet software to organize intermediate findings and format the final answer. 

(5) \textbf{Submission:} Upon completion, the final answer is saved as a \texttt{CSV} file, and the extension exports the search trajectory as a \texttt{JSON} log. Finally, we employ a post-processing script to parse these logs into the standardized trajectory format (as shown in Appendix~\ref{app:format}).

We present the statistics of human annotation time on \ours{} in Figure~\ref{fig:minitues}.

% \begin{figure}[t]
%     \centering
%     \includegraphics[width=0.9\linewidth]{fig/minutes.pdf}
%     \caption{Statistics of human annotation time on \ours{}.}
%     \label{fig:minitues}
% \end{figure}

\subsubsection{Quality Checking}
In the final phase, the triplet of query, search trajectory, and answer is distributed to a dedicated team of verifiers. This stage involves three rigorous validation steps:

(1) Verifiers first inspect the consistency of the recorded search logs. They screen for two common issues: (a) missing initial queries, where the log fails to capture the start of the session because the annotator performs the first search before activating the tool; and (b) noise injection, where irrelevant queries or browsing behaviors are recorded because the annotator fails to terminate the session immediately after the task. Trajectories exhibiting these issues are flagged for re-annotation.

(2) Upon clearing the trajectory check, verifiers evaluate the correctness of the final answer. They are granted the autonomy to conduct independent web searches, referencing the recorded trajectory at their discretion, to validate the facts. Beyond factual accuracy, verifiers strictly enforce compliance with the pre-defined format (\eg{}, specific table schema) and sorting constraints. If errors are identified (\eg{}, citation failures, missing entries, or hallucinations), the answer undergoes revision. A critical constraint is applied during this correction: the revised answer must be fully deducible from the information present in the \textit{original} search trajectory. If the original trajectory is insufficient to support the corrected answer, the instance is deemed invalid and discarded, triggering a re-annotation of the query.

\begin{table}[t]
    \centering
    \small
    \caption{Statistics of \ours{}. The input length is measured by the average number of tokens, while the output length is measured by the average number of items.}
    \begin{tabular}{crrrr}
    \toprule
        \textbf{Answer Type} & \textbf{\# Stable} & \textbf{\# Live} & \textbf{Input Len} & \textbf{Output Len} \\
    \midrule
        Item & 12 & 10 & 42.96 & 1.00 \\
        Set & 33 & 17 & 43.40 & 16.90 \\
        List & 30 & 18 & 54.23 & 13.67 \\
        Table & 148 & 105 & 98.45 & 45.98 \\
    \bottomrule
    \end{tabular}
    \label{tab:statistics}
\end{table}

Finally, to ensure that the benchmark evaluates \textit{search} capabilities rather than the \textit{parametric memory} of LLMs, we conduct a memorization check using the \texttt{DeepSeek-V3.2} model~\citep{deepseek-v3.2}. We feed the queries into the model with its reasoning and web search capabilities disabled. If the model can answer a query perfectly using only its internal knowledge, the query is considered ``solved'' by current pre-training data and is consequently excluded from the dataset. This ensures that \ours{} focuses exclusively on tasks requiring external information retrieval. After filtering, we obtain a final set of 373 queries, with detailed statistics reported in Table~\ref{tab:statistics}.

Throughout the annotation process, we recruit 15 graduate students specializing in information retrieval as expert annotators. Their domain knowledge ensures proficiency in formulating effective search strategies and synthesizing information from multiple sources. The entire pipeline, from initial question design to final answer validation, requires over one hour per query on average, reflecting the complexity and rigor of our annotation process. Figure~\ref{fig:minitues} provides detailed statistics of the time spent on the human annotation stage (\ie{}, trajectory collection and answer labeling).

\subsection{Evaluation}
\subsubsection{Data Preprocessing}
To facilitate automated evaluation, we incorporate specific output format constraints within the agent's instructions, requiring the final results to be encapsulated in \texttt{TSV} format within Markdown code blocks. We employ regular expressions to target and extract content from these blocks in the agent's final response. To ensure parsing robustness, the extraction pipeline automatically filters out redundant empty lines and handles cases where the code block may be missing by treating the entire response as raw content for fallback processing. 

Both the extracted predictions and the ground-truth answers undergo a uniform normalization procedure to eliminate the impact of formatting discrepancies. This process begins with standardizing column headers through lowercase conversion and whitespace removal. For cell values, we implement a multi-tiered cleaning logic: numeric entries are stripped of currency symbols, commas, and percentage signs, with the latter being converted to decimal equivalents. These values are then represented as strings, where integers are formatted directly and floating-point numbers are rounded to a maximum of six decimal places with trailing zeros removed. Additionally, string-based entries are normalized to lowercase and stripped of interior spaces, newlines, and Markdown artifacts such as asterisks. Finally, all null or missing values are represented as empty strings, ensuring a consistent and objective comparison at the cell level.

\subsubsection{Metrics}
Since our benchmark contains various types of answers~(\ie{}, item, set, list, and table), we employ specific metrics tailored to each type to comprehensively evaluate the performance. 
% All metrics are summarized in Table~\ref{tab:metrics}.

\noindent\textbf{Universal Metric: Exact Match}\quad First, we apply exact match (EM) as the strictest evaluation metric across all answer types. This metric assigns a score of 1 only if the generated result is completely identical to the ground truth; otherwise, the score is 0. 
% This ensures high precision in evaluating the correctness of the answer.

\noindent\textbf{Metrics for Specific Answer Types}\quad Depending on the answer type, we use additional metrics to capture different aspects of the model's performance:

(1)~\textbf{Set-type:} For answers where the output is a collection of items but the order does not matter, we use the \textbf{F1 score} to evaluate the overlap between the predicted set and the ground-truth set, allowing us to assess both precision and recall.

(2)~\textbf{List-type:} For ordered lists, both the content accuracy and the sequence of items are crucial. We initially consider using ranking-biased overlap (RBO)~\citep{rbo} to evaluate ranking quality. However, RBO cannot properly handle cases where the generated results contain duplicate items, which occasionally occurs in our task. Therefore, we adopt a two-metric approach that separately evaluates content accuracy and order accuracy.
For content accuracy, we still use the standard F1 score, which measures the precision and recall of items in the predicted list compared to the ground truth, regardless of their order. For order accuracy, we employ the Sequence Matcher algorithm, which computes an order-aware similarity score.\footnote{difflib.SequenceMatcher, \url{https://docs.python.org/3/library/difflib.html}} Given a ground-truth list and a predicted list, the order score is calculated as: $\text{Order Score} = {2M}/{T}$, 
where $M$ represents the number of matching elements (considering both content and position), and $T$ is the total number of elements in both lists. The matching process identifies the longest contiguous matching subsequences and recursively applies to remaining portions. The order score ranges from 0 to 1, where 1 indicates perfect match in both content and order. 
% This two-metric approach allows us to separately assess whether the model retrieves the correct items (F1) and whether it ranks them appropriately (Order Score), providing a comprehensive evaluation of list-type answer quality.

(3)~\textbf{Table-type:} Following the protocols of previous wide search benchmarks~\citep{widesearch}, we evaluate tabular data at two granularities. We use \textbf{row-level F1} to measure the accuracy of entire rows and \textbf{item-level F1} to assess the correctness of individual cell values.

\section{Experiments}
\subsection{Baseline Methods \& Settings}
We evaluate two paradigms of search agents on \ours{}: \textbf{ReAct-based LLM Agents} and \textbf{Commercial Agent Systems}. To ensure fair comparison, all agents receive prompts containing identical task descriptions, formatting constraints, current date information, and target queries. ReAct-based agents are additionally provided with tool-use specifications following the standard ReAct~\citep{react} workflow.

\textbf{(1) ReAct-based LLM Agents.}
We equip these agents with two core tools: \texttt{Search} and \texttt{Browse}, powered by the Google Serper API and Jina API, respectively.\footnote{Serper: \url{https://serper.dev/}, Jina: \url{https://jina.ai/reader}} Following previous studies~\citep{WebSailor, WebThinker}, content retrieved via \texttt{Browse} is summarized before being returned to the model to minimize noise. For this summarization task, we use the same underlying LLM as the agent, operating in non-thinking mode (when applicable). 
To leverage the inherent capabilities of each model, we utilize their native function-calling interfaces to construct the agents. For models supporting thinking mode, we employ an interleaved reasoning approach, allowing the model to perform multiple tool calls within a continuous thinking path. We set a maximum of 30 tool invocations per task, 8,192 output tokens per step, and disable parallel function calling for fair comparison. A retry strategy is implemented to ensure execution stability.

Our evaluation covers a diverse suite of state-of-the-art models: \textbf{Qwen3-235B-A22B}~\citep{qwen3}, \textbf{Claude 4.5 Sonnet}, \textbf{Gemini 3 Pro}, \textbf{GPT-5.2}, \textbf{DeepSeek-V3.2}~\citep{deepseek-v3.2}, \textbf{GLM-4.7}~\citep{glm-4.7}, \textbf{Seed-1.8}, \textbf{Qwen3-Max}~\citep{qwen3}, and \textbf{Kimi K2.5}. Unless otherwise specified, the native thinking mode is enabled by default with the maximum thinking budget allocated. We utilize official APIs for DeepSeek-V3.2, Qwen-Max, and Kimi K2.5, while all other models are accessed via the OpenRouter platform.\footnote{OpenRouter: \url{https://openrouter.ai/}}

\textbf{(2) Commercial Agent Systems.}
In addition to custom implementations, we evaluate closed-source commercial systems to establish a baseline for industrial-grade performance. These include \textbf{DeepSearch systems} (GPT-4o Search Preview, Perplexity Sonar Pro Search, and Google Search AI Mode) and \textbf{DeepResearch systems} (OpenAI o4-mini DeepResearch). Google Search AI Mode results are manually collected and converted to CSV format; all other systems are accessed via OpenRouter APIs.

More implementation details can be found at Appendix~\ref{app:imple}.

\begin{table*}[t]
    \centering
    \small
    \caption{Experimental results of all methods on \ours{}. The overall score is calculated by the weighted average of all EM scores. Tool calling number with $^*$ is provided by the API service provider.}
    \setlength{\tabcolsep}{.6mm}{
    \begin{tabular}{lcccccccccccc}
    \toprule
        \multirow{2}{*}[-0.5ex]{\textbf{Model / System}} & \textbf{Item} & \multicolumn{2}{c}{\textbf{Set}} & \multicolumn{3}{c}{\textbf{List}} & \multicolumn{3}{c}{\textbf{Table}} & \textbf{Overall} & \multicolumn{2}{c}{\textbf{Avg. \# Tool Calls}}  \\
        \cmidrule(lr){2-2}\cmidrule(lr){3-4}\cmidrule(lr){5-7}\cmidrule(lr){8-10}\cmidrule(lr){11-11}\cmidrule(lr){12-13}
        & \textbf{EM} & \textbf{EM} & \textbf{F1} & \textbf{EM} & \textbf{F1} & \textbf{Order} & \textbf{EM} & \textbf{Row-F1} & \textbf{Item-F1} & \textbf{EM} & \textbf{Search} & \textbf{Browse} \\
        \midrule
        \rowcolor[RGB]{235,245,250}\multicolumn{13}{c}{\textit{LLM-based ReAct Agents}} \\
        Qwen3-235B-A22B (thinking) & 40.91 & 18.00 & 52.37 & 14.58 & 36.48 & 35.96 & 4.35 & 28.32 & 43.93 & 9.65 & 2.16 & 4.03 \\
        Claude 4.5 Sonnet (non-thinking) & 59.09 & 26.00 & 60.87 & 22.92 & 58.76 & 57.78 & 9.49 & 47.85 & 63.71 & 16.36 & 10.11 & 5.67 \\
        Claude 4.5 Sonnet (thinking) & {63.64} & {28.00} & \textbf{64.86} & 22.92 & 59.24 & 56.42 & \textbf{13.04} & \textbf{49.92} & 65.17 & \textbf{19.30} & 7.57 & 4.63 \\
        Gemini 3 Pro (low) & 45.45 & {28.00} & 63.82 & \textbf{27.08} & 57.55 & 56.37 & 7.11 & 45.93 & 64.93 & 14.74 & 9.13 & 6.69 \\
        Gemini 3 Pro (high) & 50.00 & 22.00 & 62.66 & \textbf{27.08} & 60.87 & 60.12 & 8.70 & 47.01 & {66.02} & 15.28 & 11.87 & 5.79 \\
        GPT-5.2 (thinking) & 63.64 & 26.00 & 62.70 & 16.67 & 54.11 & 53.17 & 9.49 & 43.04 & 60.20 & 15.82 & 8.14 & 13.78 \\
        DeepSeek-V3.2 (non-thinking) & 22.73 & 20.00 & 52.00 & 22.92 & 56.02 & 55.45 & 6.72 & 44.14 & 62.24 & 11.53 & 12.62 & 10.18 \\
        DeepSeek-V3.2 (thinking) & 63.64 & 28.00 & 60.79 & 20.83 & {62.25} & {60.41} & 6.32 & 43.44 & 62.42 & 14.47 & 12.14 & 12.35 \\
        GLM-4.7 (thinking) & 50.00 & 22.00 & 59.44 & 20.83 & 51.99 & 50.97 & 8.30 & 43.97 & 61.28 & 14.21 & 15.75 & 12.26 \\
        Seed-1.8 (thinking) & 45.45 & \textbf{32.00} & 56.77 & 16.67 & 56.11 & 53.54 & 6.32 & 38.49 & 57.13 & 13.40 & 7.94 & 4.14 \\
        Qwen3-Max (thinking) & 59.09 & {30.00} & 63.45 & 25.00 & \textbf{66.51} & \textbf{64.08} & 10.67 & 48.48 & \textbf{66.86} & 17.96 & 11.50 & 8.94 \\
        Kimi K2.5 (thinking) & \textbf{68.18} & {28.00} & 61.71 & 18.75 & 50.52 & 48.81 & 7.91 & 45.19 & 61.23 & 15.55 & 13.02 & 7.89 \\
        \midrule
        \rowcolor[RGB]{235,245,250}\multicolumn{13}{c}{\textit{Commercial Agent Systems}} \\
        GPT-4o Search Preview & 13.64 & 4.00 & 38.70 & 8.33 & 36.65 & 36.00 & 4.74 & 29.59 & 45.61 & 5.63 & 1.00$^*$ & - \\
        OpenAI o4 Mini Deep Research  & 18.18 & 14.00 & 63.03 & 18.75 & 53.72 & 52.59 & 3.56 & 36.78 & 56.47 & 7.78 & 69.88$^*$ & - \\
        % Gemini DeepResearch & & & & & & & & & \\
        Perplexity Sonar Pro Search & 22.73 & 20.00 & 47.04 & 6.25 & 34.74 & 33.16 & 3.95 & 34.76 & 49.05 & 7.51 & - & - \\
        % Tongyi-DeepResearch & & & & & & & & & \\
        % 16 & Qwen-3-Max Search Agent \\
        Google Search AI Mode & 31.82 & 20.00 & 46.34 & 8.33 & 40.64 & 39.36 & 5.53 & 31.15 & 50.79 & 9.38 & - & - \\
    \bottomrule
    \end{tabular}
    }
    \label{tab:main_results}
\end{table*}

\subsection{Main Results}
Table~\ref{tab:main_results} presents the performance of various LLM-based agents and commercial search systems on \ours{}. We highlight several key findings from our evaluation.

\textbf{(1) Significant room for improvement remains.} Even the best-performing model, Claude 4.5 Sonnet (thinking), achieves only 19.30\% overall EM score, indicating that current search agents are far from solving complex information-seeking tasks reliably. Through manual inspection of search trajectories, we identify several recurring failure modes: (a) limited problem decomposition capabilities, where agents fail to devise effective search plans for complex queries; (b) insufficient self-correction abilities, where agents struggle to adjust strategies based on intermediate search results; and (c) inadequate web traversal skills, where agents fail to exploit hyperlinks within pages, leading to incomplete information gathering.

\textbf{(2) Task complexity scales with information breadth.} Performance degrades substantially as the required information scope increases. Agents perform reasonably well on item-type questions but struggle significantly on table-type tasks, which demand collecting and organizing information across multiple dimensions. Notably, in \ours{}, the amount of information to be gathered does not necessarily correlate with question difficulty, \ie{}, an item-type question may still require extensive search before arriving at the correct answer. The pronounced performance drop on structured answer formats (lists and tables) suggests that current models face challenges not only in information collection but also in answer organization and formatting.

\textbf{(3) Efficient tool usage matters.} Interestingly, the best-performing Claude 4.5 Sonnet model uses tools quite efficiently, with moderate numbers of search and browse calls compared to other models. In contrast, models like DeepSeek-V3.2 and GLM-4.7 invoke substantially more tools yet achieve lower scores. This suggests that excessive tool usage does not necessarily improve performance; rather, noise from irrelevant retrieved content and increased context length may negatively impact reasoning quality.

\textbf{(4) Reasoning mode provides consistent gains.} Comparing thinking and non-thinking variants of the same model family reveals clear benefits from extended inference-time computation. Claude 4.5 Sonnet improves from 16.36\% to 19.30\% overall EM, and DeepSeek-V3.2 improves from 11.53\% to 14.47\% when thinking mode is enabled. However, these gains come at the cost of significantly higher token consumption, presenting a trade-off between performance and efficiency.

\textbf{(5) Commercial search systems underperform.} Surprisingly, commercial deep research and search products do not outperform LLM-based ReAct agents. GPT-4o Search Preview, which performs only a single retrieval per query, is clearly insufficient for the complex tasks in \ours{}. OpenAI o4 Mini Deep Research and Perplexity Sonar Pro Search, despite their sophisticated pipelines, suffer from poor instruction-following, resulting in numerous formatting errors in their final answers. Google Search AI Mode demonstrates relatively better performance among commercial systems; while it lags behind LLM-based agents in accuracy, it offers substantial advantages in response speed, positioning itself closer to traditional search engine experiences.

\subsection{Further Analysis}
We conduct further analyses to better understand model behavior and identify potential directions for improvement.

\noindent\textbf{Comparison with Human Search Behavior.}
To better understand the gap between model and human search strategies, we compare the behavioral patterns of Claude 4.5 Sonnet (thinking) with human annotations, analyzing the average number of search queries and browsed pages, query refinement patterns, and the correlation between human-model similarity and task performance (detailed metrics for all models are provided in Appendix~\ref{app:behavior}). We observe that humans and models exhibit different strategies: humans issue fewer queries (3.53 on average) but browse substantially more pages (19.03), whereas models search more frequently (7.57 queries) but browse far fewer pages (4.63). This suggests that humans favor thorough exploration within search results, while models rely on repeated querying rather than deep examination of retrieved content. Additionally, humans show higher adjacent query overlap (0.31 vs. 0.22), indicating more targeted query refinement, whereas models tend to construct less related queries between consecutive searches. We also find a positive correlation between human-model behavioral similarity and task performance: the high-similarity group achieves an average F1 of 0.76, compared to 0.56 for the low-similarity group, and successfully solved cases exhibit higher URL overlap rates (0.31 vs. 0.15). These findings suggest that aligning model behavior more closely with human search strategies, particularly in terms of deeper content exploration and more coherent query refinement, may be a promising direction for improving agent performance. 

\begin{figure}[t]
     \centering
     \begin{minipage}[b]{0.49\linewidth}
         \centering
         \includegraphics[width=\linewidth]{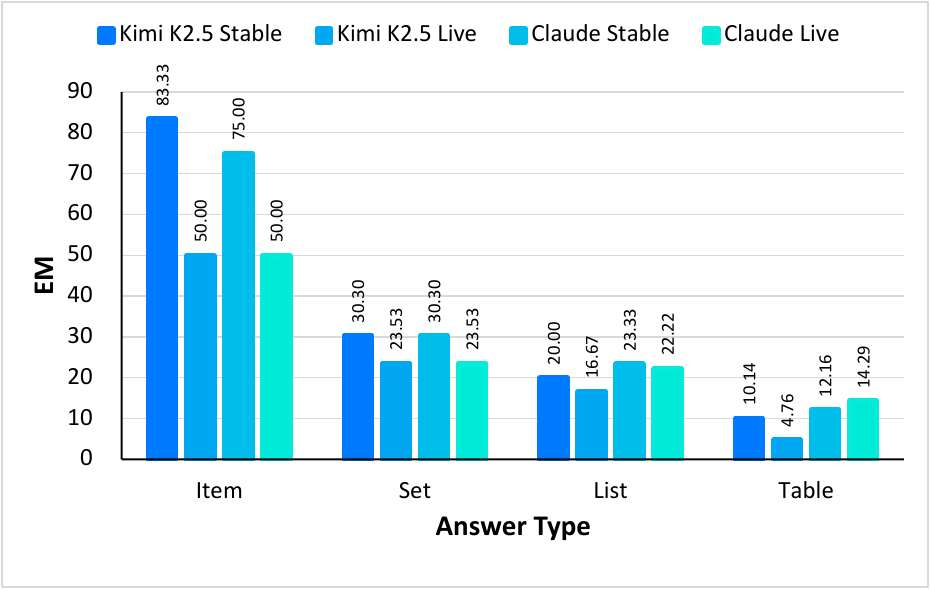}
         \caption{Performance of Kimi K2.5 (thinking) and Claude 4.5 (thinking) on different subsets of \ours{}.}
         \label{fig:live}
     \end{minipage}
     \hfill
     \begin{minipage}[b]{0.49\linewidth}
         \centering
         \includegraphics[width=\linewidth]{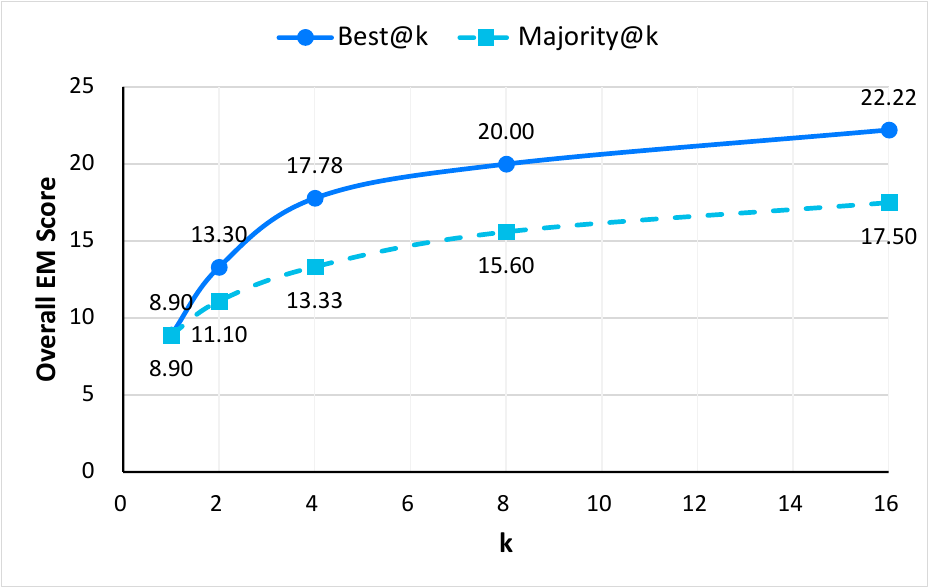}
         \caption{Inference-time scaling performance of Qwen3-Max over 40 random samples.}
         \label{fig:scaling}
     \end{minipage}
\end{figure}

% \begin{figure}
%     \centering
%     \includegraphics[width=.9\linewidth]{fig/live.pdf}
%     \caption{Performance of Kimi K2.5 (thinking) and Claude 4.5 (thinking) on different subsets of \ours{}.}
%     \label{fig:live}
% \end{figure}

\noindent\textbf{Performance Difference across Question Types.}
We analyze model performance across stable and live subsets to explore the potential impact of data contamination. Intriguingly, Kimi K2.5, the most recently released (\textit{Jan. 28, 2026}) model in our evaluation, achieves notably lower performance on the live subset compared to the stable subset (11.33\% vs. 18.39\% overall EM), while other models~(Claude 4.5 Sonnet) show no significant difference. We hypothesize that this difference arises because Kimi K2.5's training data, being the most recent, is more likely to contain answers to stable questions. In contrast, older models may not have memorized either subset, making their performance uniformly dependent on search capabilities. This finding validates the design of our live subset: as models are trained on increasingly current data, static benchmarks risk measuring memorization rather than search ability. The live subset, with its periodically updated answers, provides a more robust evaluation that remains resistant to such contamination over time.

% \begin{figure}
%     \centering
%     \includegraphics[width=0.9\linewidth]{fig/scaling.pdf}
%     \caption{Inference-time scaling performance. }
%     \label{fig:scaling}
% \end{figure}
\noindent\textbf{Inference-time Scaling.}
Recent research has shown that allocating more computational resources at inference time can effectively improve model performance~\citep{deepseek-v3.2,widesearch,browsecomp}. To investigate whether search agents benefit from such scaling, we conduct experiments on 40 randomly sampled queries using Qwen3-Max, generating $k$ independent runs per query ($k \in \{1, 2, 4, 8, 16\}$). We report Best@$k$ (whether any attempt succeeds) and Majority@$k$ (confidence-weighted voting). As shown in Figure~\ref{fig:scaling}, both metrics improve consistently as $k$ increases. Best@$k$ rises from 8.90\% to 22.22\% at $k$=16, a 2.5$\times$ improvement, suggesting that models have potential capabilities not reliably activated in single attempts. Majority@$k$ also improves but consistently lags behind Best@$k$, indicating that selecting the correct answer from multiple candidates remains challenging. These results demonstrate that inference-time scaling offers a promising direction for improving search agents, while also highlighting the need for better answer verification mechanisms.

\begin{wrapfigure}{r}{.48\linewidth}
    \centering
    \includegraphics[width=\linewidth]{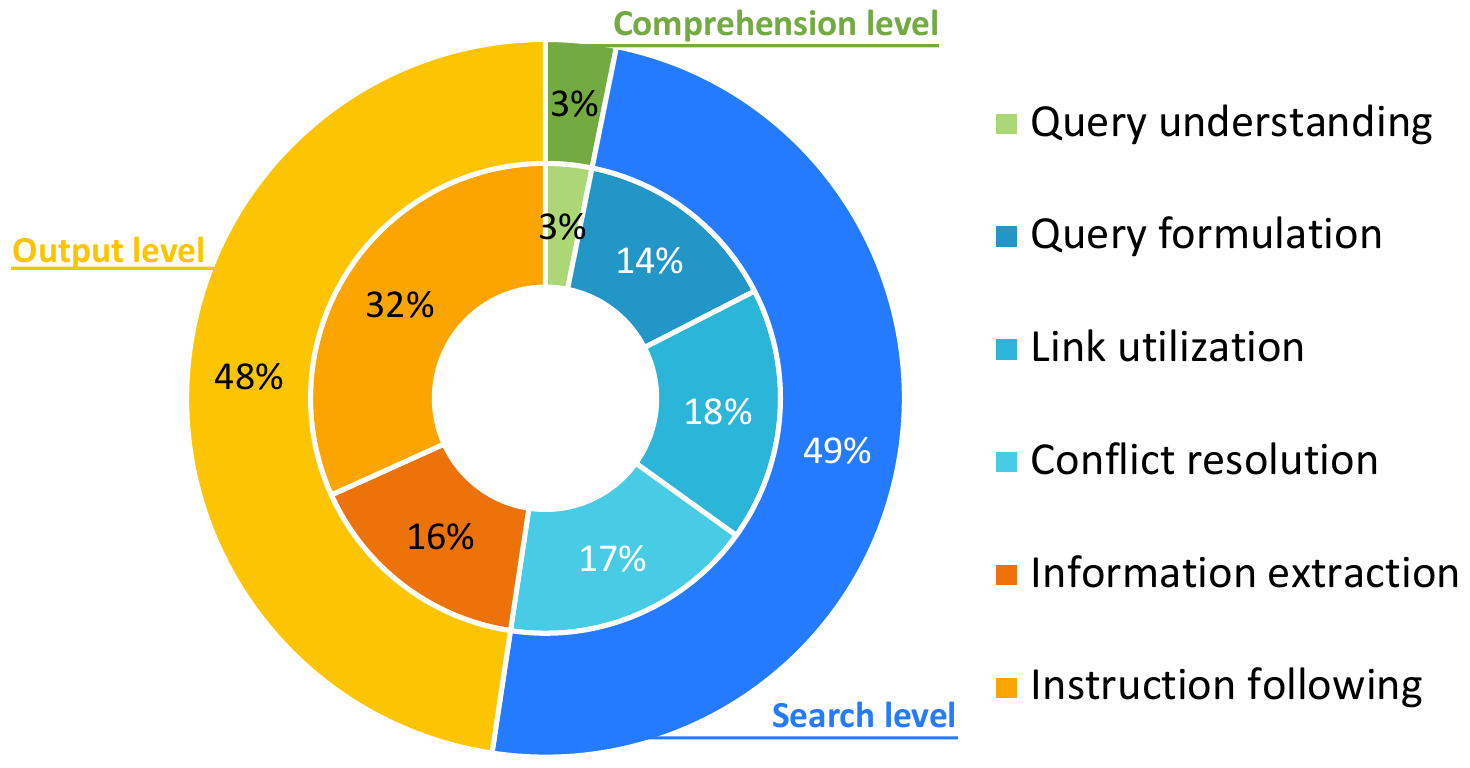}
    \caption{Illustration of different error types and their ratios.}
    \label{fig:error}
\end{wrapfigure}
\noindent\textbf{Error Analysis.}
To gain deeper insight into the failure modes of current search agents, we manually analyze 50 error cases from the best-performing model (Claude 4.5 Sonnet with thinking). As shown in Figure~\ref{fig:error}, we categorize the identified errors into three levels based on where they occur in the search pipeline (noting that a single sample may exhibit multiple error types).
(1) At the \textbf{comprehension level}, query misunderstanding accounts for only 3.2\% of errors, indicating that current LLMs possess strong semantic understanding capabilities. (2) The majority of failures occur at the \textbf{search level}, comprising 49.2\% of all errors. These include inability to formulate effective queries (14.3\%), failure to exploit hyperlinks within pages for deeper exploration (17.5\%), and inability to initiate verification queries when encountering conflicting information across sources (17.5\%). These errors suggests that current agents lack effective strategies for thorough web exploration and information verification. (3) At the \textbf{output level}, we observe failures in accurate information extraction from retrieved pages (15.9\%) and instruction-following errors (31.7\%), such as generating incorrect table headers. Notably, instruction-following errors constitute the largest single category, which is consistent with our main finding that commercial systems often struggle with output formatting.

\begin{table*}[t]
    \centering
    \small
    \caption{Average cost per question for each model on \ours{}. The input and output token counts reflect only the main reasoning chain, while tokens consumed by the browsing tool are reported separately. Prices are reported in US dollars per million tokens, with RMB converted at a rate of 7:1.}
    \setlength{\tabcolsep}{.9mm}{
    \begin{tabular}{lccccccc}
    \toprule
        \textbf{Model} & \textbf{\# Input} & \textbf{\# Output} & \textbf{\# Tool Input} & \textbf{\# Tool Output} & \textbf{Input Price} & \textbf{Output Price} & \textbf{Average Cost} \\
    \midrule
        Qwen3-235B-A22B (thinking) & 162888.63 & 46399.86 & 134491.03 & 32838.69 & 0.29 & 2.86 & 0.31 \\
        Claude 4.5 Sonnet (non-thinking) & 321680.56 & 7185.19 & 156620.54 & 5173.38 & 3.00 & 15.00 & 1.62 \\
        Claude 4.5 Sonnet (thinking) & 267248.29 & 7619.70 & 133077.07 & 4308.98 & 3.00 & 15.00 & 1.37 \\
        Gemini 3 Pro (low) & 242960.98 & 27886.18 & 134920.32 & 6702.21 & 2.00 & 12.00 & 1.17 \\
        Gemini 3 Pro (high) & 292792.10 & 35536.54 & 138258.97 & 7145.17 & 2.00 & 12.00 & 1.37 \\
        GPT-5.2 (thinking) & 471289.43 & 20408.00 & 234744.65 & 14354.61 & 1.75 & 14.00 & 1.72 \\
        DeepSeek-V3.2 (non-thinking) & 499634.20 & 11204.50 & 222052.20 & 7938.16 & 0.29 & 0.43 & 0.22\\
        DeepSeek-V3.2 (thinking) & 594672.30 & 16671.11 & 256626.42 & 10004.80 & 0.29 & 0.43 & 0.26\\
        GLM-4.7 (thinking) & 567243.06 & 45994.95 & 278159.63 & 41877.75 & 0.29 & 1.14 & 0.34 \\
        Seed-1.8 (thinking) & 358530.71 & 15270.81 & 216757.47 & 10905.33 & 0.11 & 1.14 & 0.10 \\
        Qwen3-Max (thinking) & 444898.78 & 15354.54 & 199051.86 & 10042.04 & 0.36 & 1.43 & 0.27 \\
        Kimi K2.5 (thinking) & 416011.44 & 14367.39 & 183547.81 & 10318.11 & 0.57 & 3.00 & 0.42 \\
    \bottomrule
    \end{tabular}
    }
    \label{tab:cost}
\end{table*}

\noindent\textbf{Cost Analysis.}
Table~\ref{tab:cost} presents the average token consumption and cost per query for each model evaluated on \ours{}. We report the number of input and output tokens in the main reasoning chain, tokens consumed by the browsing tool, official API prices, and the resulting average cost per query. Based on the results, we have several observations. 
First, Claude 4.5 Sonnet, despite achieving the best performance, requires relatively high costs (\$1.37--\$1.62 per query) due to its premium pricing (\$3.00/\$15.00 per million input/output tokens). However, Claude demonstrates remarkable efficiency in token usage, consuming significantly fewer tokens than most other models. Interestingly, the thinking mode variant is both more effective and cheaper than its non-thinking counterpart, as it uses tools more efficiently and generates fewer total tokens. Second, Chinese models (DeepSeek, GLM, Seed, Qwen, and Kimi) offer substantially lower costs, ranging from \$0.10 to \$0.42 per query. Among these, Seed-1.8 achieves the lowest cost at \$0.10 per query, while DeepSeek-V3.2 provides an attractive balance between cost (\$0.26) and performance. Third, we observe that token consumption does not necessarily correlate with performance. Models like DeepSeek-V3.2 and GLM-4.7 consume the most tokens (over 500K input tokens per query) yet do not achieve top performance, suggesting that effective tool usage and reasoning quality matter more than sheer volume of computation. Finally, GPT-5.2 incurs the highest average cost (\$1.72) due to both high token consumption and relatively expensive pricing, without delivering proportionally better results. These findings highlight the importance of considering cost-effectiveness when deploying search agents in practice, as the most expensive models are not always the best performing.

\section{Limitations}
While \ours{} provides a comprehensive benchmark for evaluating information-seeking agents, we acknowledge several limitations.
First, due to the complexity of the annotation process, which requires substantial human involvement at every stage, the scale of our benchmark remains relatively limited (373 queries). Although sufficient for evaluation purposes, this size may not support large-scale model training such as supervised fine-tuning.
Second, \ours{} currently focuses exclusively on text-based information seeking and does not incorporate multimodal content such as images or videos. As search agents evolve to integrate visual capabilities from GUI agents, future benchmarks should consider evaluating agents' ability to process and extract information from multimodal web content, which would better simulate real-world human information gathering.
Third, due to cost constraints, we set a maximum of 30 tool invocations per query in our experiments. While this limit is sufficient for most tasks, we observed that a small number of cases failed to collect adequate information due to this constraint. Future work could explore more flexible resource allocation strategies.

\section{Ethical Considerations}
We strictly adhere to ethical standards throughout the construction of \ours{}. For annotation tool development, our browser extension is engineered to record \textit{only} task-relevant interactions within specific search sessions, with no personally identifiable information, cookies, or extra browsing history collected. All annotators participated voluntarily with informed consent. To mitigate potential biases, we intentionally diversified query domains, though we acknowledge that the benchmark may still underrepresent certain languages or specialized domains. Regarding potential misuse, \ours{} focuses exclusively on publicly available information and does not include queries related to sensitive personal data or harmful content.

\section{Conclusion}
This paper introduces \ours{}, a benchmark for evaluating general information-seeking agents, comprising 373 human-crafted queries with four structured answer formats (item, set, list, and table). \ours{} addresses key limitations of existing benchmarks by unifying the evaluation of deep reasoning and broad information aggregation, enabling deterministic evaluation through structured answers, incorporating a live subset with periodically updated answers to resist data contamination, and providing complete human search trajectories for process-level supervision. Our evaluation of mainstream LLMs and commercial search products reveals that current systems still fall substantially short of human performance, with even the best model achieving only 18.23\% overall exact match. These results highlight significant room for improvement in problem decomposition, adaptive search planning, and efficient tool utilization, and we hope \ours{} will serve as a valuable resource for developing more capable information-seeking agents.

\section{Acknowledgments}
This work was supported by the National Natural Science Foundation of China (Grant No. 62402497) and the China Postdoctoral Science Foundation under Grant Number 2025T180440. The authors would like to thank Jinghan Yang, Zhao Wang, Binyu Xie, Shangze Li, Yifei Chen, Zhixin Lin, and Yuyao Zhang for data annotations.

\appendix
\begin{figure}[t]
    \centering
    \includegraphics[width=.6\linewidth]{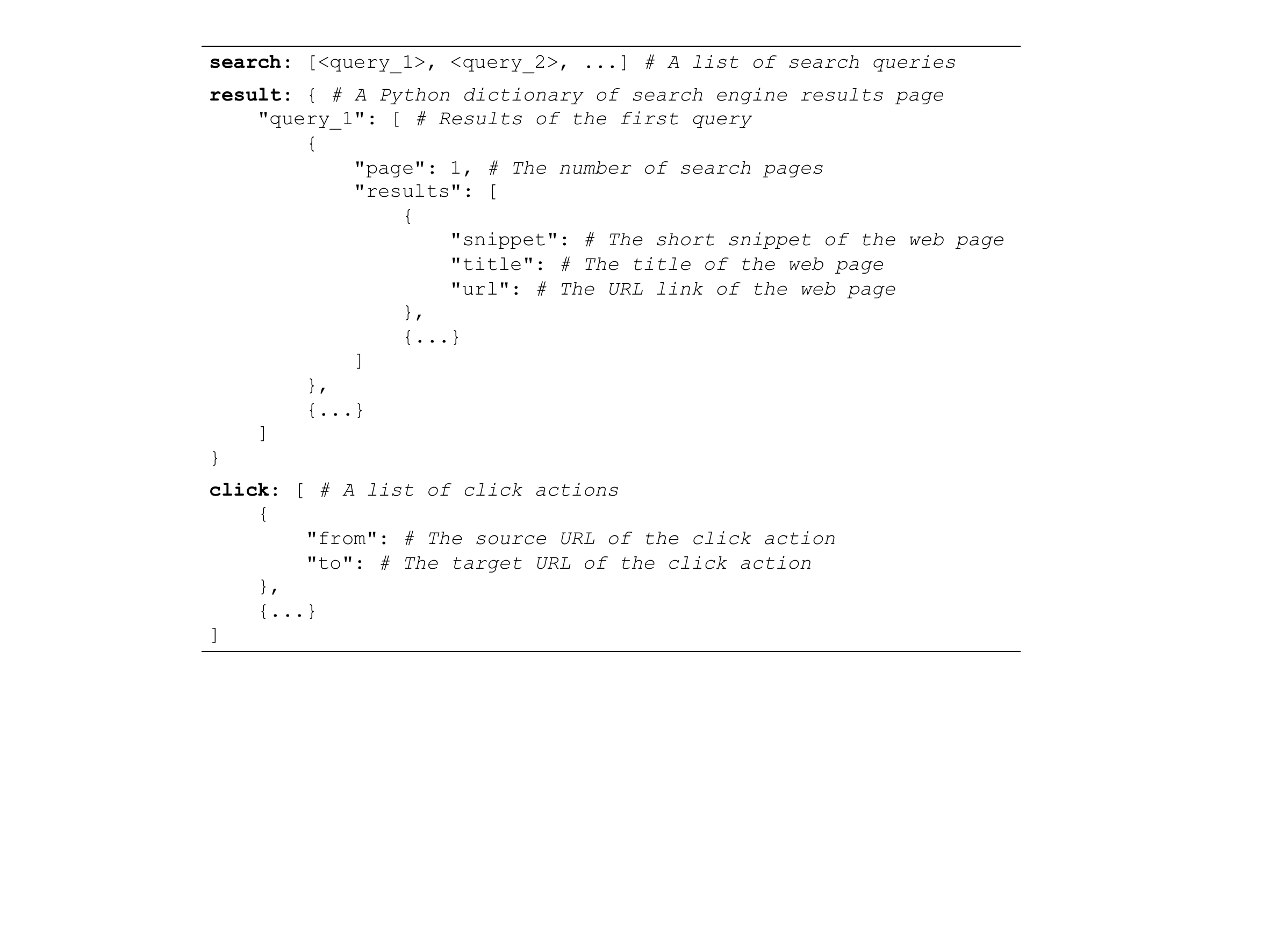}
    \caption{The \texttt{JSON} template of annotated trajectories.}
    \label{fig:trajectory_example}
\end{figure}
\section{Data Format}\label{app:format}
\ours{} is organized into three components: questions, answers, and human trajectories.

(1)~\textbf{Questions:} Each question is stored as a \texttt{JSON} object containing the unique identifier, encrypted question text, answer type, question type (stable or live), and topic category. The encryption is applied to prevent data contamination from model pre-training.

(2)~\textbf{Answers:} Each answer is stored as a separate \texttt{CSV} file corresponding to its question identifier. The \texttt{CSV} format accommodates all four answer types: item answers are placed in the first cell, set and list answers contain a single column, and table answers contain multiple columns with a predefined schema.

(3)~\textbf{Human Trajectories:} Each human trajectory is stored as a \texttt{JSON} file containing three fields: the search queries issued by the annotator, the search engine results for each query, and the click actions recording navigations between pages. An example is illustrated in Figure~\ref{fig:trajectory_example}.

\begin{table*}[t]
    \centering
    \small
    \caption{Comparison of search behaviors between different models and human annotators.}
    \begin{tabular}{lccc}
    \toprule
        \textbf{Model} & \textbf{Search Similarity} & \textbf{Search Diversity} & \textbf{Browsing Similarity}  \\
    \midrule
        Qwen3-235B-A22B (thinking) & 30.59 & 43.05 & 13.08 \\
        Claude 4.5 Sonnet (non-thinking) & 32.21 & 24.03 & 17.83 \\
        Claude 4.5 Sonnet (thinking) & 30.02 & 21.73 & 17.63 \\
        Gemini 3 Pro (low) & 35.98 & 38.36 & 14.14 \\
        Gemini 3 Pro (high) & 36.30 & 36.54 & 14.12\\
        GPT-5.2 (thinking) & 24.82 & 23.00 & 8.81 \\
        DeepSeek-V3.2 (non-thinking) & 35.54 & 26.74 & 13.42 \\
        DeepSeek-V3.2 (thinking) & 38.97 & 28.87 & 13.82 \\
        GLM-4.7 (thinking) & 34.36 & 29.00 & 14.92 \\
        Seed-1.8 (thinking) & 31.42 & 39.43 & 17.42 \\
        Qwen3-Max (thinking) & 36.68 & 31.81 & 15.69 \\
        Kimi K2.5 (thinking) & 32.69 & 25.92 & 16.82 \\
    \bottomrule
    \end{tabular}
    \label{tab:behavior_metrics}
\end{table*}

\section{Implementation Details}\label{app:imple}
Following recent studies~\citep{WebSailor, WebThinker,DeepSearchAgentSurvey}, we adopt the ReAct~\citep{react} architecture for our agent implementation. The system prompt and tool definitions are provided in Figure~\ref{fig:react_prompt} and Figure~\ref{fig:tools_def}, respectively. Each model is equipped with two tools: \textit{Search} and \textit{Browse}, with a maximum limit of 30 tool invocations per session.

\begin{itemize}[topsep=0pt,leftmargin=2em]
    \item \textbf{Search Tool:} This tool is implemented using the Google Search engine (through Serper API). It accepts a query as input and retrieves the top-10 search results.
    \item \textbf{Browse Tool:} This tool is built upon the Jina API to extract content from a given URL. To reduce noise, the retrieved content is summarized by an LLM. For experimental consistency, the summarization model is identical to the agent's backbone model, using the non-thinking mode to reduce cost.
\end{itemize}
The prompt for commercial systems follows a similar structure and is shown in Figure~\ref{fig:base_prompt}.

\begin{figure}[t]
\centering
\begin{promptbox}
You are a deep research assistant. Your core function is to conduct thorough, multi-source investigations into any topic. You must handle both broad, open-domain inquiries and queries within specialized academic fields. For every request, synthesize information from credible, diverse sources.
You have 30 chances to call tools, use them wisely.

\textbf{\# Final Answer Format}\\
When you have gathered sufficient information, you must output the final answer within \texttt{<answer></answer>} tags. Inside these tags, you must strictly follow the \textbf{TSV (Tab-Separated Values)} format enclosed in a code block \texttt{\textasciigrave\textasciigrave\textasciigrave tsv}.

Determine the nature of the answer (Item, List, or Table) and format it as follows:

\begin{enumerate}[leftmargin=*, noitemsep]
    \item \textbf{Single Item (Fact/Value)}: Use a single column with the header \texttt{Value}.
    \item \textbf{List}: Use a single column with the header \texttt{Item}.
    \item \textbf{Table (Structured Data)}: Use standard TSV with appropriate headers for each column.
\end{enumerate}

\textbf{CRITICAL}:
\begin{itemize}[leftmargin=*, noitemsep]
    \item The content inside \texttt{\textasciigrave\textasciigrave\textasciigrave tsv} must be valid TSV format.
    \item Always include a header row.
    \item Do not add markdown notes or explanations \textit{inside} the code block. Put any summary text \textit{outside} the code block but still inside the \texttt{<answer>} tags.
\end{itemize}

Current date: \{current\_date\}\\
User Question: \{question\}
\end{promptbox}
\caption{The system prompt used in ReAct-based Agent.}
\label{fig:react_prompt}
\end{figure}

\begin{figure}[t]
\centering
\begin{promptbox}[title=\textbf{Prompt for Commercial Systems}] 
You are a helpful assistant. Given an user's question, your task is to thinking step by step and output the final answer in the format of TSV.

\textbf{\# Final Answer Format}\\
You must output the final answer within \texttt{<answer></answer>} tags.
Inside these tags, you must strictly follow the \textbf{TSV (Tab-Separated Values)} format enclosed in a code block \texttt{\textasciigrave\textasciigrave\textasciigrave tsv}.

Determine the nature of the answer (Item, List, or Table) and format it as follows:

\begin{enumerate}[leftmargin=*, noitemsep]
    \item \textbf{Single Item (Fact/Value)}: Use a single column with the header \texttt{Value}.
    \item \textbf{List}: Use a single column with the header \texttt{Item}.
    \item \textbf{Table (Structured Data)}: Use standard TSV with appropriate headers for each column.
\end{enumerate}

\textbf{CRITICAL}:
\begin{itemize}[leftmargin=*, noitemsep]
    \item The content inside \texttt{\textasciigrave\textasciigrave\textasciigrave tsv} must be valid TSV.
    \item Always include a header row.
    \item Do not add markdown notes or explanations \textit{inside} the code block. Put any summary text \textit{outside} the code block but still inside the \texttt{<answer>} tags.
\end{itemize}

Current date: \{current\_date\}\\
User Question: \{question\}
\end{promptbox}
\caption{The system prompt used in commercial systems.}
\label{fig:base_prompt}
\end{figure}

\begin{figure}[t]
\centering
\begin{toolbox}
% \vspace{0.5em}
\textbf{1. Search}
\begin{itemize}[leftmargin=1.5em, noitemsep]
    \item \textbf{Description}: Perform a Google web search and return the top search results.
    \item \textbf{Parameters}:
    \begin{itemize}[leftmargin=1em]
        \item \param{query}{string, required}: The search query string to be issued to Google.
    \end{itemize}
\end{itemize}

\vspace{0.5em}
\textbf{2. Browse}
\begin{itemize}[leftmargin=1.5em, noitemsep]
    \item \textbf{Description}: Visit one or more web pages and return a summarized version of their content based on a specific goal.
    \item \textbf{Parameters}:
    \begin{itemize}[leftmargin=1em]
        \item \param{url}{array of strings, required}: One or more URLs of the web pages to visit.
        \item \param{goal}{string, required}: The specific information objective to focus on when summarizing the web pages.
    \end{itemize}
\end{itemize}

\end{toolbox}
\caption{The functional schema of the tools.}
\label{fig:tools_def}
\end{figure}

\section{Human-Model Search Behavior Analysis}\label{app:behavior}
We provide detailed metrics quantifying the similarity between model and human search behaviors. We compute three metrics for each model, which are defined as:
\begin{itemize}[topsep=0pt,leftmargin=2em]
    \item \textbf{Search Similarity} measures the overlap between model-generated queries and human queries. For each human query, we compute the Jaccard similarity with all model queries and take the maximum value. The Jaccard similarity is defined as the ratio of the intersection to the union of term sets: $Sim(Q_h, Q_m) = |T_h \cap T_m| / |T_h \cup T_m|$, where $T_h$ and $T_m$ are the term sets of human and model queries after lowercasing and tokenization. The question-level similarity is the average of all human query scores, and the final score is averaged across all questions.
    \item \textbf{Search Diversity} measures the diversity of consecutive queries, computed as the average Jaccard similarity between adjacent queries. Lower values indicate more diverse query reformulations.
    \item \textbf{Browsing Similarity} measures the overlap between URLs visited by the model and those visited by human annotators, computed as the Jaccard similarity of the two URL sets.
\end{itemize}

From the results shown in Table~\ref{tab:behavior_metrics}, we can observe: First, search similarity scores range from 24.82\% (GPT-5.2) to 38.97\% (DeepSeek-V3.2 thinking), indicating moderate alignment between model and human query formulations. Second, Claude 4.5 Sonnet exhibits the lowest search diversity (21.73\%), suggesting more focused and incremental query refinement similar to human behavior, while Qwen3-235B-A22B shows the highest diversity (43.05\%), indicating more exploratory search patterns. Third, browsing similarity remains relatively low across all models (8.81\%--17.83\%), confirming that models and humans often follow different navigation paths even when seeking the same information. Notably, Claude 4.5 Sonnet achieves the highest browsing similarity (17.83\%), which aligns with its superior task performance.

\end{document}